\documentclass[conference]{IEEEtran}
\ifCLASSINFOpdf
\else
\fi
\usepackage{graphicx}

\begin{document}
%
\title{Masked Face Recognition using ResNet-50}

\author{\IEEEauthorblockN{Bishwas Mandal}
\IEEEauthorblockA{Kansas State University\\
Manhattan, KS, USA\\
Email: bishmdl76@ksu.edu}
\and
\IEEEauthorblockN{Adaeze Okeukwu}
\IEEEauthorblockA{Kansas State University\\
Manhattan,KS, USA\\
Email: adaeze@ksu.edu}
\and
\IEEEauthorblockN{Yihong Theis}
\IEEEauthorblockA{Kansas State University\\Manhattan, KS, USA\\Email:yihong@ksu.edu}}


%


\maketitle

\begin{abstract}
Over the last twenty years, there have seen several outbreaks of different coronavirus diseases across the world. These outbreaks often led to respiratory tract diseases and have proved to be fatal sometimes. Currently, we are facing an elusive health crisis with the emergence of COVID-19 disease of the coronavirus family. One of the modes of transmission of COVID-19 is airborne transmission. This transmission occurs as humans breathe in the droplets released by an infected person through breathing, speaking, singing, coughing, or sneezing. Hence, public health officials have mandated the use of face masks which can reduce disease transmission by 65\%. For face recognition programs, commonly used for security verification purposes, the use of face mask presents an arduous challenge since these programs were typically trained with human faces devoid of masks but now due to the onset of Covid-19 pandemic, they are forced to identify faces with masks. Hence, this paper investigates the same problem by developing a deep learning based model capable of accurately identifying people with face-masks. In this paper, the authors train a ResNet-50 based architecture that performs well at recognizing masked faces. The outcome of this study could be seamlessly integrated into existing face recognition programs that are designed to detect faces for security verification purposes.  
\end{abstract}


%
\IEEEpeerreviewmaketitle

\section{Introduction}
As humans continue to make technological advances, the need to secure devices that house both our private and official matters is paramount. Some of the traditional methods achieved this feat by using ID cards, passwords, passphrases, and puzzles. However, with the rapid advancement in the field of deep learning and high performance computing, the usage of human biometrics such as the face, voice and fingerprints are deemed ubiquitous in the modern day security verification programs\cite{minaee_biometric_2020}. The widespread use of these human biometrics relates to their uniqueness, making it difficult to replicate them\cite{han_study_2005}. Similarly, face recognition programs allow a quicker yet efficient framework for identification of an individual\cite{petrescu_face_2019}. 

Face recognition software can be seen in everyday devices like mobile phones and laptops and in physical security devices deployed in offices. Their success in accurately identifying different people is unprecedented. However, to achieve this step, pre-trained models such as FaceNet\cite{schroff_facenet_2015}, ResNets\cite{he2015deep}, SesNets, and their variants in use in human face recognition are trained on human faces without a mask. Generally, this is not an issue because people verify their identity without facial coverings. However, in situations where there’s a need to identify someone with facial covering, it becomes challenging for deep learning algorithms to identify the unique features that verify a person's identity. 

Currently, we face a critical health challenge with the onset of COVID-19 disease that has infected about 51 million people and killed over 1.3 million people worldwide\cite{noauthor_global_2020}. Public health officials identified that wearing a facial mask can reduce transmission of COVID-19 by about 65\%\cite{chu_physical_2020}. Abiding by this public health directive, individuals are always required to wear a face mask, especially at times when they interact with one another. Therefore, individuals in enclosed spaces will need to wear face masks to authenticate their identity on their mobile phones or laptop devices. Currently, algorithms that are successful on unmasked faces have been unable to generalize such successes on masked faces. One of the problems associated to detecting an unmasked face is that the deep learning models would use more features, i.e., the whole face to identify someone. However, with a masked face, the nose and mouth is occluded. In that lies the problem of identifying individuals with just the eyes and sometimes, the forehead. 

Hence, we proposed a framework to solve the problem of identifying individuals with masked faces using ResNet-50 architecture. Furthermore, we apply transfer learning to adapt a pre-trained ResNet-50 model to our images, comprising of individuals without face masks. Then we apply different architectural changes and subject the model to hyperparameter tuning to identify the identity of individuals with a mask from images of same individuals without a mask.

Our contribution includes:
\begin{enumerate}
\item Finetuning a pre-trained ResNet-50 model on our dataset and obtaining an accuracy of 89\%.
\item Developing and finetuning the hyperparameters of a ResNet-50 based architecture that gives around 47\% percent accuracy in identifying faces when used on the masked face dataset.
\item A detailed description on hyperparameters tuning for the model is presented.
\end{enumerate}


\section{Related Work}
Computer vision is one of the most successful research areas in the field of machine learning. Major progress occurred in this area within the past two decades. In 2006, Hinton et al \cite{hinton_fast_2006} proposed a model to train deeper neural networks to the desired depth. Then, with the availability of ImageNet \cite{deng_imagenet_nodate}, a breakthrough occurred in 2012 \cite{krizhevsky_imagenet_nodate}, with the development of an 8-layer neural network that outperformed all other algorithms in the identification of images in ImageNet. They applied convolutional layers, rectified linear activation, and dropout in their architecture. Their results paved the way for developing better algorithms in computer vision.
Face recognition can be classified as a subset of computer vision. With the application of similar techniques used in computer vision, face recognition present high accuracies in recognizing faces. This is important since it enables the identification and authentication of individuals\cite{petrescu_face_2019}.  In this section, we focus our discussion on the deep learning techniques that evolved the work on face recognition research. In 2015, FaceNet \cite{schroff_facenet_2015} , using the GoogleNet-24 framework achieved an accuracy of 99.63\% on the LFW database\cite{huang_labeled_nodate}. Parkhi et al \cite{Parkhi15} worked on the VGGFace and achieved an accuracy of 98.95\% on the LFW database. In this case, the architecture used was the VGGNet-16. Most common techniques have relied on ResNet\cite{he2015deep} and its variants after Microsoft Research performed an image classification with this architecture with a good performance in 2015. Analogous to our particular case, Cao et al\cite{cao2018vggface2} used ResNet-50 architecture to access face recognition performance in their work.  

Recognizing faces with occlusion is a variant of the facial recognition problem. Simple face recognition algorithms become limited\cite{unknown} when the intention is to recognize faces when people wear hats, eyeglasses, masks as well as other objects that can occlude part of the face while leaving others unaffected in the images. For instance, the existing methods such as the VGG-16, ResNet, and VGGFace learnt discriminate features from full-face images. However, learning features from full-face images of occluded faces does not give us useful features on the identity of the face. Needless to say, that our work is related to some of these problems. In the current study, we focused on solving the problem of facial recognition on masked faces. We realized that when a person wears a mask in the training image, it becomes cahllenging to recognize the corresponding person without a mask, since the eyes and nose are important features for identification.

Several works have been conducted to detect faces with occlusion. There are two main approaches used in this regard, which are the restoration approach and discard occlusion based approach \cite{unknown}.  The restoration approach tries to restore the occluded parts of the images based on the images in the training. Bachi et al \cite{bagchi2014robust} use a 3D face recognition system in restoring parts of the face which is occluded.  The system registers a 3D input of a person's face using the Iterative Closest Point Algorithm, then the occluded parts of the face are detected, followed by restoration technique using the Principal Component Algorithm(PCA).
 Similarly, Drira et al \cite{drira_3d_2013} use a 3D based statistical approach in recognizing and estimating the occluded part of the face. THE PCA approach is used in the recovery of occluded parts of the face.\smallskip

The discard occlusion based approach rejects the occluded parts completely to avoid a bad reconstruction process, then the remaining parts of the face are used in the feature extraction and classification processes. Priya et al \cite{priya_occlusion_2014} divided the face image into small local patches. Then used the support vector machine to detect occluded parts of the face and removed them. After which the mean based weight matrix is used to identify the remaining partial face. Also, Weng et al\cite{weng_robust_2016} eliminated occluded parts of the face, marked out keypoints, and extracted features from those keypoints. Furthermore, they used a matching mechanism to align the extracted information with those in the gallery by estimating the similarity of the two faces as the distance between to aligned features.

During the COVID-19 pandemic, lots of researchers have focused their study on whether or not people wear masks \cite{wang2020masked,loey_hybrid_2021,jiang2020retinamask}. However, this is different from our study since the problem of wearing or not wearing a mask is a mask detection problem while ours is strictly a face detection problem, with an added constraint that the face is masked. With regards to this, we realize there are far less works that are studying how to use the current state-of-the-art facial recognition models to detect the identification of a person with a face mask. Detecting the identity of a person wearing a face mask is a much tougher problem since the number of features to make this identification reduces. Hariri\cite{unknown} worked on recognizing faces on a mask by cropping the eyes and the forehead, and then used a quantization based pooling method on VGG–16 pre-trained model by considering only the feature maps at the last convolutional layer using Bag-of-Features (BoF) paradigm. Geng et. al\cite{10.1145/3394171.3413723} used two strategies to identify masked faces. First, the simulated masked faces of the plain faces was used to generate more training data. Next, they used the Domain Constrained Ranking (DCR) loss that creates two centers for the full image and masked images of each identity. Then the DCR ensures that the feature of masked faces gets close to the corresponding full face. Masked face recognition can also be considered as partial face recognition problem. Liao et al\cite{6296663} developed an alignment-free face representation method based on Mult-Keypoint Descriptors (MKD), where the variable size descriptor of a face is determined by the actual content of the image. Then used a fast filtering method for facial recognition. Sato et al\cite{670963} proposed a radial basis function (RBF) networks for 100 persons partial face images. However, He et al.\cite{He2018DynamicFL} proposed a dynamic feature match (DFM) method with sliding loss to address partial face image issues regardless of the size. 

Our work is closely related to Hariri\cite{unknown}. We use ResNet instead of VGG16 because of the high performance of CNN based methods that have strong robustness to illumination, facial expression and facial occlusion changes. In this study, we demonstrate a method to take on the mask as another feature besides the full face and deploy deep CNN based model to address the problem of masked face detection during the COVID-19 pandemic. The paper is further divided into different sections. Section III discusses the problem statement, Section IV examines the technical approach considered and implemented in this study, Section V explores the experimental approach, Section VI analyzes our preliminary results and finally, the Section VII concludes with the future research direction.

\section{Problem Statement}
Identifying the identity of masked faces is a challenging problem for computer vision models since the features required to accurately predict the identity of an individual is reduced from the whole face to just the eye and sometimes the forehead. This study is built on existing pre-trained ResNet-50 architecture trained on human faces to solve the problem of identifying a person’s identity when wearing a face mask.

\begin{figure*}[ht]
\includegraphics[width=\textwidth,height=12.5cm]{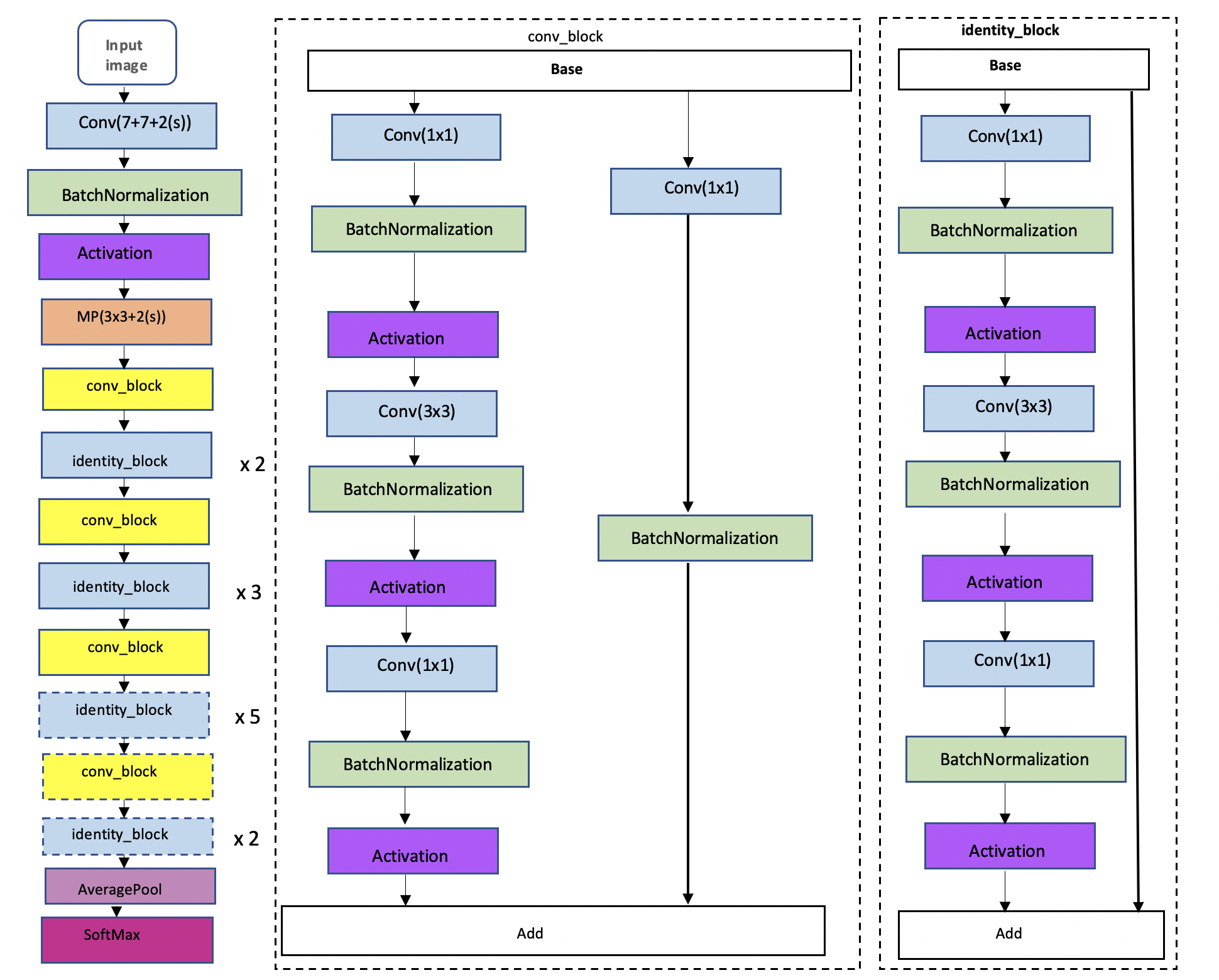}
\caption{ResNet-50 architecure }
\label{fig:resnet}
\end{figure*}

\section{Technical Approach}
In this section, we explain the methods applied to solve the proposed problem, followed by detailed description on our achievement for this work, specifically on the architecture we implemented and other techniques that we applied to get better results.

\subsection{Method}
Considering that the objective of our project is to identify individuals with a face mask, we applied a transfer learning technique where a model developed on another face recognition task will apply to our specific project. We use transfer learning on a convolutional neural network-based model, in this case, a ResNet-50 architecture. We selected ResNet-50 based on its performance in several image recognition projects, particularly, it has the best time and memory performance when compared to VGGNet19 and DenseNet121\cite{8981749}.  We use such a pre-trained model obtained from this work \cite{cao2018vggface2} and fine-tuned the parameters on our dataset of faces without masks. Next, we ran our model on the masked faces and fine-tuned the model based on our results.  The objective was to identify an individual with a mask after training our data on faces without a mask. 

We considered alternative approaches. We identified that the occlusion on the face presents a challenge to our model since there is a reduction in the number of features on the masked face that presents nonuniform feature mapping. Therefore we considered cropping the occluded part of the face on the masked face and running our algorithm with the updated features. We implemented a simple cropping feature to do this and got uneven results because of the images that had head coverings such as hats and caps. An option would have been to remove these images but since we had scarce data, we decided to drop this avenue. Another option we considered was to use a segmentation algorithm to detect masks on faces and perform cropping action but due to time consideration, this pursuit was discarded.

Also, we considered using supervised domain adaptation to the resulting model. For this purpose, we considered faces without masks as the source domain (S) and the faces with masks as the target domain (t). Then, we trained and tested the model based on two different scenarios. For the first scenario, we trained the model only on S and test the performance on T. Then after, we trained the model on S and portion of T and tested the model on the other portion of T. 

Since, we did not see lots of prior work in this area with the same dataset, we chose our baseline as the performance of our algorithm in detecting masked faces to that of the performance of our algorithm in detecting unmasked faces. We tried to compare to the state-of-the-art results in face recognition and found through literature review \cite{hu_when_2015}, that such comparisons are considered unfair since the datasets are different. 

\subsection{Transfer Learning}
Transfer learning\cite{noauthor_deep_nodate} is a method whereby a model applied to a machine learning task A is adapted and applied to another machine learning task B. Transfer learning solves the problem of insufficient training data and improves performance when modeling the target task\cite{olivas_handbook_2009}. 

In this study, we apply transfer learning by using a pre-trained model from this work \cite{cao2018vggface2} and adapt it to our model.  Choosing this model made it easy to train on our image dataset since its an architecture used for a task that is analogous to ours, which implies that the model expects images as an input. Since this model is trained on a large corpus of faces, we know that those parameters in the architecture will transfer to our data. We expect our model to efficiently learn to extract features from our dataset to predict identities accurately.

\subsection{ResNet-50 Architecture}
ResNet-50, with 50 layers is one of the variants of ResNet \cite{he2015deep}, a convolutional neural network. It has 48 Convolution layers along with 1 MaxPool and 1 Average Pool layer. Fig. 2 shows the architecture of ResNet-50 in detail.  ResNet \cite{he2015deep} is based on the deep residual learning framework. It solves the problem of the vanishing gradient problem even with extremely deep neural networks. Resnet-50, despite having 50 layers has over 23 million trainable parameters which is very much smaller than existing architectures. 

\begin{figure}[ht]
\includegraphics[width = 0.48\textwidth]{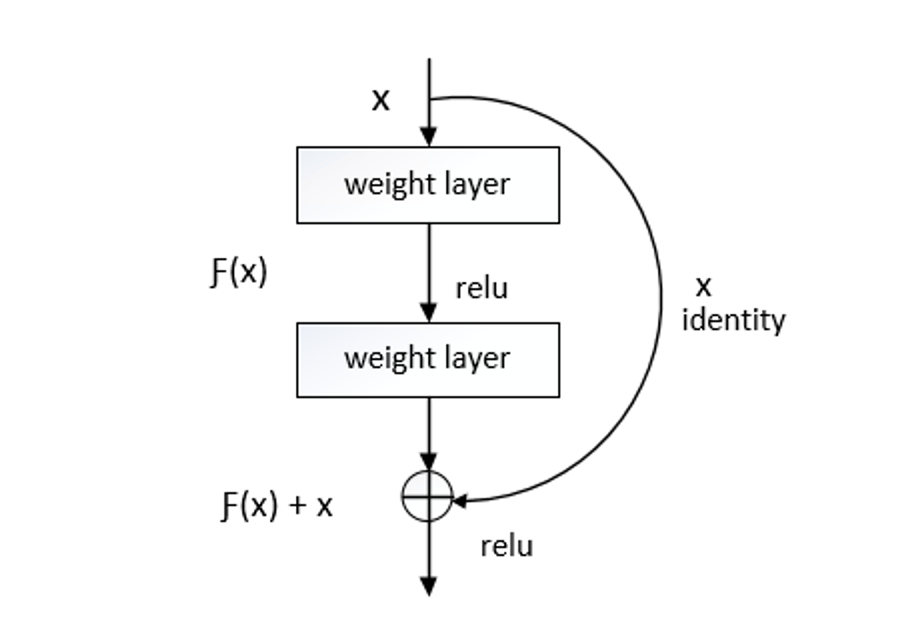}
\caption{Residual Learning block}
\label{fig:learnblock}
\end{figure}

The reasoning behind its performance is still open to discussions, but the simplest way to understand is to explain residual blocks and how these blocks work.

Let us consider a neural network block, whose input is x, where we would like to learn the true distribution H(x). Let us denote the difference (or residual) between this as:

\[R(x) = Output - Input = H(x) - x\] 

Rearranging it we get, \[H(x) = R(x) + x\] 

The residual block is trying to learn the true output, H(x). Taking a closer at the image above, we realize that since we have an identity connection coming due to x, the layers are learning the residual, R(x). The layers in a traditional network are learning true output (H(x)) while the layers in a residual network are learning the residual (R(x)). Also, it is observed that it is easier to learn the residual of the output and input rather than the input only. In this manner, the identity residual model allows for the reuse of activation functions from previous layers since they are skipped and add no complexities to the architecture.

In the proposed work, we take ResNet-50 architecture and pre-trained weights \cite{cao2018vggface2} from a model trained on ResNet-50,which was pre-trained on MS1M first, and then fine-tuned on VGGFace2.


\section{Experimental Setup}
We performed our experiment using Jupyter notebook, Pytorch v1.2.0 library on  Ubuntu version 18.04.4 LTS(Bionic Beaver)operating system, running on GP104 (GEForce GTX 1080).

\begin{figure}[ht]
\centering
  \includegraphics[height=6cm]{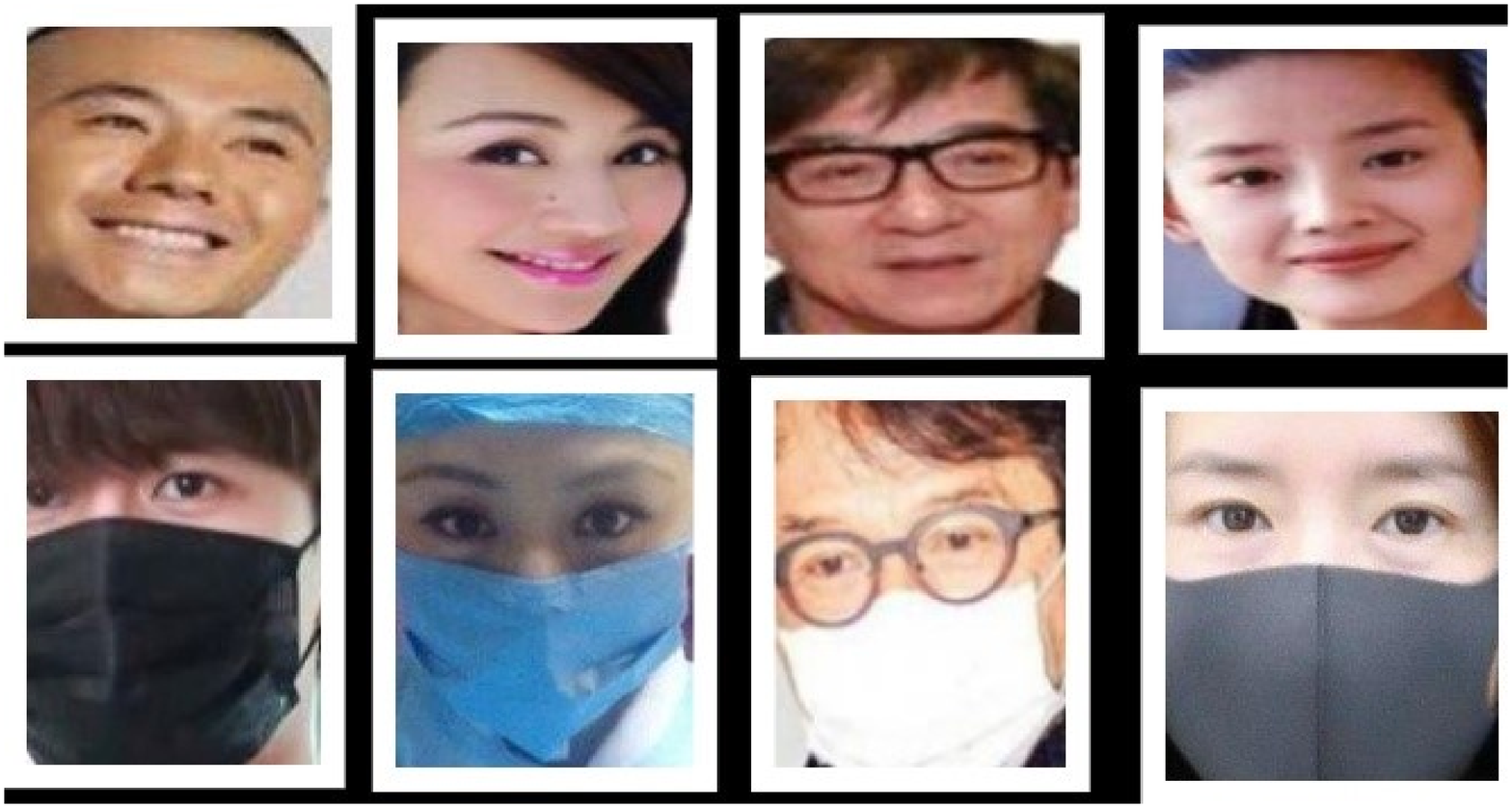}
  \caption{Pairs of face images from RMFRD dataset: face without mask (up) and face with mask (down).}
  \label{fig:collage}
\end{figure}

\begin{figure*}
\includegraphics[width=\textwidth,height=11.0cm]{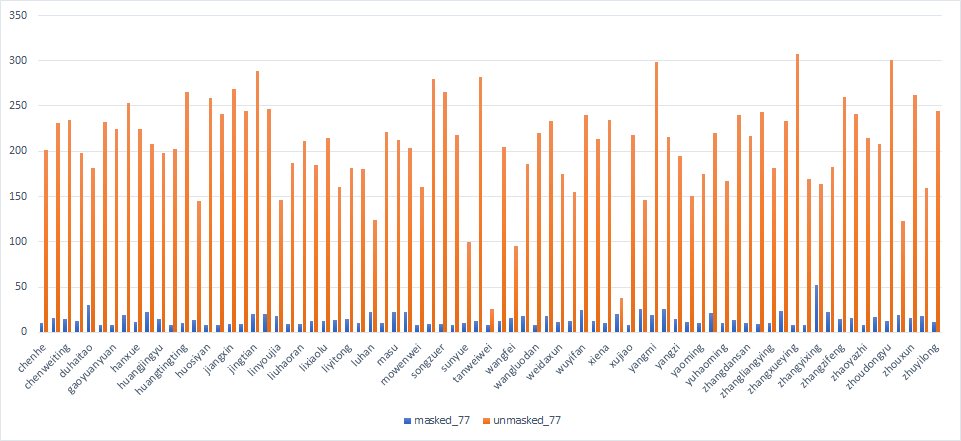}
\caption{Different Classes versus Number of Images}
\label{fig:class}
\end{figure*}

\subsection{Datasets}
We use the “Real-world masked face recognition dataset” (RMFRD) \cite{wang2020masked}, which is a masked face dataset devoted mainly to improve the recognition performance of the existing face recognition technology on the masked faces during the COVID-19 pandemic. The datasets contain three different data, real-world masked face recognition dataset, simulated masked face recognition datasets, and real-world masked face verification. In this project, we chose specifically to use the real-world masked face recognition dataset that contains 5,000 masked faces of 525 people and 90,000 unmasked faces. Figure ~\ref{fig:collage} presents some pairs of face images without a mask and with a mask for the corresponding faces.

Out of 5,000 masked and 90,000 unmasked faces, we take only the images of the classes that have at least eight images for the masked faces prediction and the same classes for the unmasked faces prediction. Both masked and unmasked dataset is split into 70\% training and 30\% validation data. With the given criterion we end up with 77 classes in the masked dataset and unmasked dataset. We made this decision because tried and tested implementations of face recognition using convolutional networks have shown that we need 10 - 20 images per class if we are using a pre-trained model and 1000-2000 images per class if the model is trained from scratch\cite{warden_how_2017}. We do not have such an enormous dataset available, therefore we had to settle for at least eight images per class in the masked dataset. In Figure ~\ref{fig:class}, we show the statistics of the dataset used in our project, where the blue bars are for the masked dataset and the orange bars are for the unmasked dataset. From the bar chart, we can see that we have an unbalanced dataset with the unmasked dataset for each class surpassing the masked dataset with an average ratio of 50:1.

The dimension of the images are not fixed but approximately between 100-200 pixels by 100-200 pixels. In our training and validation phases, we convert all the images to a size of 180 by 180 pixels, and based on the experimentation so far, using this size turns out to be the wisest of choices. We applied the random flip horizontally to our training images as a data augmentation method. We applied this to our dataset such that at each batch generation, these augmented data will be produced. With data augmentation, we think that the diversity of our dataset will increase during training. 

\subsection{HyperParameter Tuning}
We manually selected and applied different hyperparameter tuning in a bid to get better results. Since we had two different experiments on training our model on unmasked faces, then test on masked faces. During the training and testing the following hyperparameters work best for our model as shown in Table \ref{table:1}. It is important to note that from our table we changed our optimizer from Stochastic Gradient Descent(SGD) in the unmasked dataset to Adam in the masked dataset. This was because we realized that our accuracy improved from 21\% to the current 44.73\% when we did so. 

\begin{table}[ht]
\centering
\caption{Hyperparameters tested}
\begin{tabular}{ |c|c|c| } 
 \hline
  HyperParameters & unmasked & masked \\
  \hline \hline
 Batch Size & 256 & 32 \\ 
 Optimizer & SGD & Adam \\ 
 Dropout & 0.5 & 0.4 \\ 
 Loss & Cross Entropy & Cross Entropy \\ 
 Learning Rate & 0.002 & 0.0016 \\ 
 Epoch & 20 & 25 \\ 
 \hline
\end{tabular}
\label{table:1}
\end{table}

\section{Results}
In this section, we discuss our best results in terms of accuracy and loss. We will use precision, recall, and F1-score to observe how well our model worked since these are the commonly used performance metrics. Finally, we will compare our results to similar works in this area, however with different datasets to see how good our model performs.
\subsection{Unmasked Faces}
For our unmasked faces, our architecture did better on the masked than on the masked dataset. This is as a result of having more dataset for training the unmasked than the masked dataset. However, its performance is close to though lower than the performance of the state of the art algorithms in face recognition. We attribute this that we might need to do more hyperparameter tuning to get better results. Also, we could try other CNN models and compare the results to our architecture.  Also, we acknowledge that comparison of CNN models trained on different dataset results in unfair comparisons hence we do not use such as a baseline for our results\cite{hu_when_2015}.

We explain the different hyperparameters selected for our unmasked face training. We started with a Batch size of  256, SGD with momentum of 0.9, and nesterov was enabled. Since we are using transfer learning, we froze 10 children of the pre-trained network and trained the rest. We use a Learning Rate of  0.002 with a decay rate (gamma) of 0.11 and a step size of 14. We run our proposed model for 20 epochs with a cross-entropy loss. We noticed that our model was overfitting, hence we added dropout layers for regularisation of the model. We achieved an accuracy of 89.7016\% and a loss of  0.4698 (refer to Fig. 4 and Fig. 5). Additionally we got, a precision score of 0.8993, a recall score of 0.8970, and F1 score of 0.897 as shown in Table \ref{table:nonlin}.\medskip

\begin{figure}
\includegraphics[width = 0.48\textwidth]{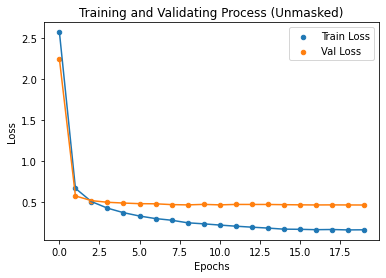}
\caption{Training and Validation Loss for Unmasked model}
\label{fig:umasked_loss}
\end{figure}

\begin{figure}
\includegraphics[width = 0.48\textwidth]{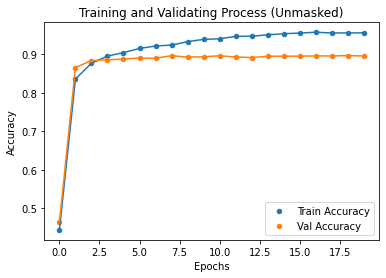}
\caption{Training and Validation Accuracy for Unmasked model}
\label{fig:unmasked_acc}
\end{figure}

\subsection{Masked Faces}
 After we have trained our model on our training sets, we tested on our masked data, and we realized that we did not do as well as we did in the unmasked face recognition.  This is because of the lower number of features due to the occlusion of the face by the mask. The data imbalance of our unmasked to the masked dataset can also be a problem. We compare the two results in Table \ref{table:nonlin} below. To achieve these results, we tuned several hyperparameters. We changed our batch size from 256 to a batch size of  32. We used the optimizer, Adam, instead of Stochastic Gradient Descent. Then we froze seven children of the proposed model. We use the learning rate of  0.0016 with a decay rate (gamma) of 0.1 and a step size of 14. We run our algorithm for 25 epochs with cross-entropy loss, and we achieved an accuracy of 47.91\% and a loss of  2.4092. Time taken for the training process was 1m 48s. Since we experienced overfitting in the training set, we added dropout layers for regularisation. Our performance metrics are precision of 0.4613, recall of 0.4719, and F1 score of 0.4473.

\subsection{Comparison of Masked and Unmasked Results}
\begin{figure}
\includegraphics[width = 0.48\textwidth]{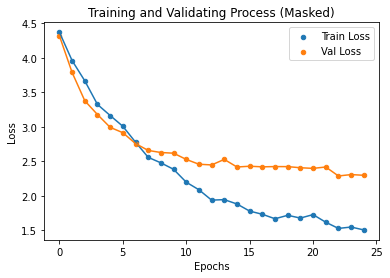}
\caption{Training and Validation Loss for Masked model}
\label{fig:loss_masked}
\end{figure}

When we compare the training and validation loss of the unmasked dataset to that of the masked dataset in Figure \ref{fig:umasked_loss} and \ref{fig:loss_masked}, we can see that the unmasked dataset gives us the perfect loss curve that shows that our model is learning from the data. However, for the masked dataset, we noticed that the training curve does not reach a minimum.  Although we see the validation curve quickly gets to a minimum and becomes stable. But the problem is that the validation loss is too high, which implies that our model is not learning from our data, and we need to do more work to get it to perform better. 

\begin{figure}
\includegraphics[width = 0.48\textwidth]{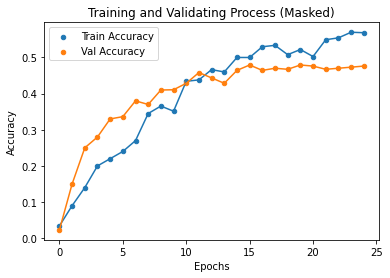}
\caption{Training and Validation Accuracy for Masked model}
\label{fig:acc_masked}
\end{figure}

As expected, we have a nice curve for our training and validation accuracy for the unmasked dataset, as shown in Figure \ref{fig:unmasked_acc}. We can still do better to increase the accuracy of the validation dataset. We compare this result to that of the masked dataset in Figure \ref{fig:acc_masked} and notice that our train and validation accuracy keeps increasing without giving that stable curve at a point. 
This result corresponds to our expectation earlier in our report that the reduction in features due to occlusion may hinder our masked model from learning. Hence we can try other techniques to improve the performance of our face recognition on masked faces. We will discuss such methods in the next section. Finally, looking at our performance metrics, we observe that our model performs well on the unmasked than on the masked. The F1-score for the masked and unmasked is 0.897 and 0.4473. The recall was also high for the unmasked model while it was low for the masked model. We also have similar results for precision.

\begin{table}[ht]
\caption{Performance Metrics}
\centering 
\begin{tabular}{c c c c} 
\hline\hline 
Type of Data & Precision & Recall & F1-Score\\ [0.5ex] 
\hline 
unmasked & 0.8933 & 0.8970 & 0.897 \\ 
masked & 0.4613 & 0.4719 & 0.4473 \\ [1ex] 
\hline 
\end{tabular}
\label{table:nonlin} 
\end{table}

\section{Conclusions and Future Work}
In the course of this work, we have learned that convolutional networks are good at recognizing faces but can perform badly when faces are occluded. The implication is that the simple use of these architectures will not give the desired outcome as we noticed in our project of recognizing masked faces. Hence, more novel techniques need to be implemented to improve their performance. 
We have seen from literature that cropping out the masked part of the face can give a better result, or in the case of 3D learning, where the full faces were recovered from the structural features available and then features compared to training images. We can use improve on this method to improve our results. We also learned we could try increasing the dataset available to us by simulating masks on our unmasked dataset. We think this could give us a better result as this will remove the unbalanced nature of our data. Also, we could try techniques such as domain adaptation methods to see how this would boost our performance. Finally, we could also combine other CNN architectures and also combine these architectures with other machine learning techniques to also improve our performance.




%

\bibliographystyle{unsrt}
\bibliography{reference,maskface1,MFaces2}


\end{document}